\documentclass[10pt, a4paper]{article}
\usepackage{amsmath, mathtools}
\usepackage{subcaption}
\DeclareMathOperator*{\argmax}{argmax}
\usepackage{lrec-coling2024} 
\usepackage{ragged2e,tabularx, makecell}
\usepackage{float}
\usepackage{enumitem}
\usepackage{graphicx}
\usepackage{adjustbox}
\usepackage{makecell} 
\usepackage{multirow}
\usepackage{hanging}

\title{Automatic Alignment of Discourse Relations of Different Discourse Annotation Frameworks 
\\ \vspace*{.5\baselineskip}}

\name{Yingxue Fu} 

\address{School of Computer Science, University of St Andrews, Scotland, UK, KY16 9SX \\
         yf30@st-andrews.ac.uk\\}

\abstract{
 Existing discourse corpora are annotated based on different frameworks, which show significant dissimilarities in definitions of arguments and relations and structural constraints. Despite surface differences, these frameworks share basic understandings of discourse relations. The relationship between these frameworks has been an open research question, especially the correlation between relation inventories utilized in different frameworks. Better understanding of this question is helpful for integrating discourse theories and enabling interoperability of discourse corpora annotated under different frameworks. However, studies that explore correlations between discourse relation inventories are hindered by different criteria of discourse segmentation, and expert knowledge and manual examination are typically needed. Some semi-automatic methods have been proposed, but they rely on corpora annotated in multiple frameworks in parallel. In this paper, we introduce a fully automatic approach to address the challenges. Specifically, we extend the label-anchored contrastive learning method introduced by~\citet{zhang-etal-2022-label} to learn label embeddings during discourse relation classification. These embeddings are then utilized to map discourse relations from different frameworks. We show experimental results on RST-DT~\citep{carlson-etal-2001-building} and PDTB 3.0~\citep{prasad-etal-2018-discourse}.
 \\ \newline  
 }

\begin{document}

\maketitleabstract

\section{Introduction}
Discourse relations are an important means for achieving coherence. Previous studies have shown the benefits of incorporating discourse relations in downstream tasks, such as sentiment analysis~\citep{wang-etal-2012-exploiting}, text summarization~\citep{huang-kurohashi-2021-extractive} and machine comprehension~\citep{narasimhan-barzilay-2015-machine}. Automatic discourse relation classification is an indispensable part of discourse parsing, which is performed under some formalisms, the notable examples including the Rhetorical Structure Theory (RST)~\citep{mann1988rhetorical}, based on which the RST Discourse Treebank (RST-DT) is created~\citep{marcu1996building}, and a lexicalized Tree-Adjoining Grammar for discourse (D-LTAG)~\citep{WEBBER2004751}, which forms the theoretical foundation for the currently largest human-annotated discourse corpus\textemdash the Penn Discourse Treebank (PDTB)~\citep{Prasad2006ThePD, prasad-etal-2018-discourse}\footnote{
We focus on RST and PDTB because our method requires a large amount of data and these two frameworks have been applied to the annotation of corpora that overlap in selected texts, thus mitigating the effect of domain shift in the results. Our method does not require corpora built on the same texts.}.  

As discourse annotation has a high demand on knowledge about discourse, discourse corpora are costly to create. However, these discourse formalisms typically share similar understanding of discourse relations and their role in discourse construction. Thus, an option to enlarge discourse corpora is to align the existing discourse corpora so that they can be used jointly. This line of work starts as early as~\citet{hovy1992parsimonious}, but it remains challenging to uncover the relationship between discourse relations used in different annotation frameworks. 

\begin{figure*}[h!]
\centering
\includegraphics[width=0.7\textwidth, height=0.3\textheight, scale=1.0]{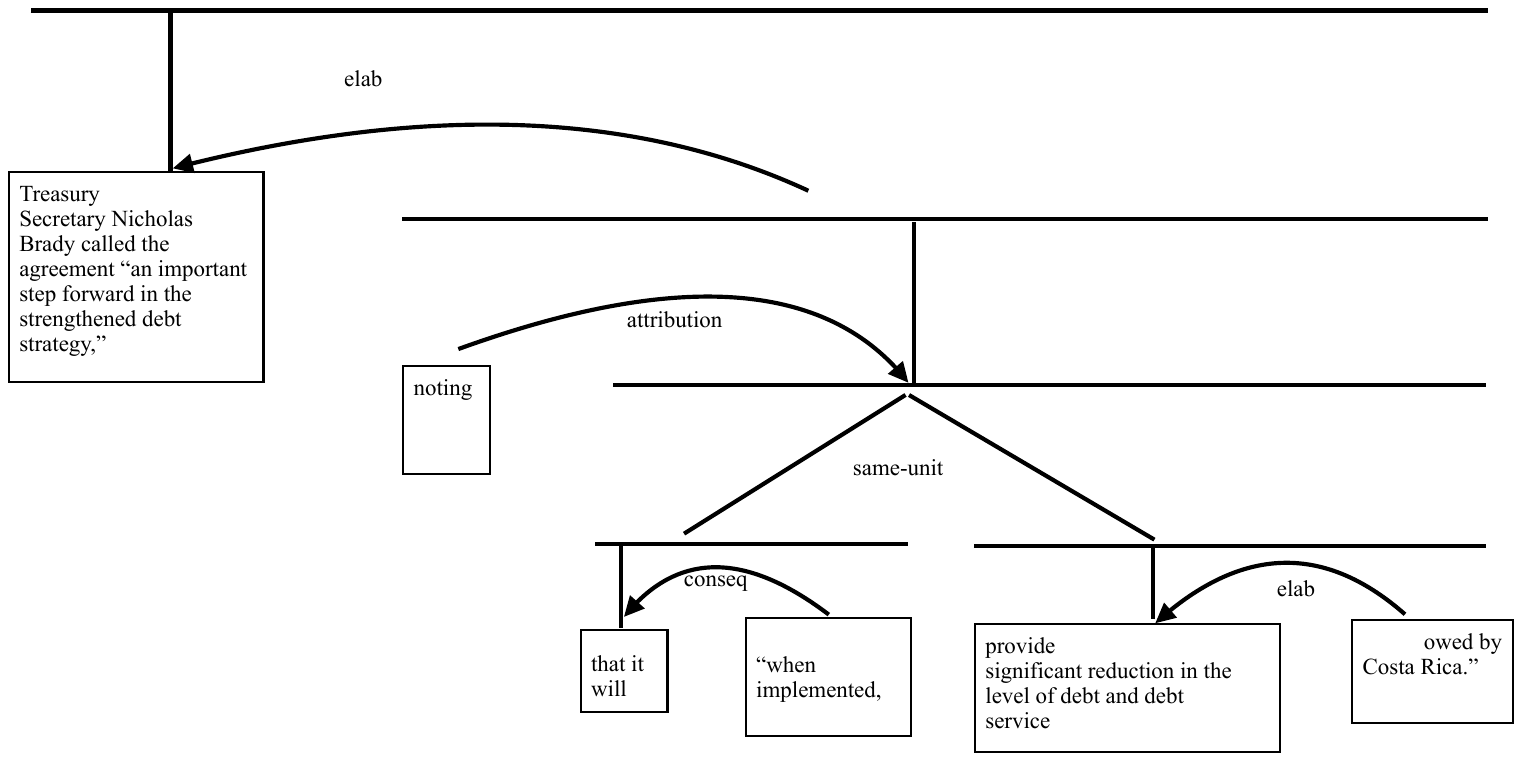}
 \vspace{-1\baselineskip}
  \caption{RST-style annotation (wsj\_0624 in RST-DT).}
  \label{rst-wsj0624}
\end{figure*}

Figure~\ref{rst-wsj0624} shows an example of RST-style annotation. The textual spans in boxes are EDUs and the arrow-headed lines represent asymmetric discourse relations, pointing from satellites to nuclei. The labels \textit{elab(oration)} and \textit{attribution} denote discourse relations. As the two spans connected by the relation~\textit{same-unit} are equally salient, the relation is represented by undirected parallel lines. The spans are linked recursively until a full-coverage of the whole text is formed, as shown by the uppermost horizontal line. The vertical bars highlight the nuclei.

As RST-DT and PDTB have an overlapping section of annotated texts, the corresponding PDTB-style annotation on the same text is:

\begin{enumerate}[before=\normalfont\footnotesize]
 \item \textit{the agreement ``an important step forward in the strengthened debt strategy''}, ~\textbf{that it will ``when implemented, provide significant reduction in the level of debt and debt service owed by Costa Rica.''}
 (implicit,  given, Contingency.Cause.Reason)

 \item \textit{that it will  provide significant reduction in the level of debt and debt service owed by Costa Rica.}, ~\textbf{implemented,}	
 (explicit, when, Temporal.Asynchronous.Succession)

 \item \textit{that it will  provide significant reduction in the level of debt and debt service owed by Costa Rica.}, ~\textbf{implemented,}	
 (explicit, when, Contingency.Cause.Reason)
\end{enumerate}

\noindent where Argument 1 (Arg1) is shown in italics and Argument 2 (Arg2) is in bold. The annotations in parentheses represent \textit{relation type}, which can be implicit, explicit or others, \textit{connective}, which is identified or inferred by annotators to signal the relation, and \textit{sense label}, which is delimited with dots, with the first entry showing the sense label at level 1 (L1 sense), the second entry being the sense label at level 2 (L2 sense) and so on.

The task presents a challenge owing to a multitude of factors. First, different formalisms have distinctive assumptions about higher-level structures and discourse units. PDTB focuses on semantic relations between arguments, and argument identification is performed following the \textit{Minimality Principle}, which means that only those parts that are necessary and minimally required for understanding a relation are annotated~\citep{prasad-etal-2008-penn}. In comparison, elementary discourse units (EDUs) in RST are typically clauses. It has been shown repeatedly that segmentation criteria affect the scope of discourse relations and influence the type of relations that can be attached~\citep{demberg2019compatible, benamara-taboada-2015-mapping,rehbein-etal-2016-annotating}. 

In the first annotation of PDTB, Arg1, i.e., \textit{the agreement ``an important step forward in the strengthened debt strategy''}, is taken from the original text ``Treasury Secretary Nicholas Brady called the agreement ``an important step forward in the strengthened debt strategy'''' and the part ``Treasury Secretary Nicholas Brady called'' is not covered because it does not contribute to the interpretation of the relation here. In contrast, this part is kept in an EDU in RST. 

Another major difference between the two frameworks is that RST enforces a tree structure, and all the EDUs and complex discourse units (CDUs) (spans formed by adjacent elementary discourse units and adjacent lower-level spans) should be connected without crossings, while PDTB only focuses on local relations without commitment to any higher-level structure, as exemplified by the three independent annotations shown above. Previous studies~\citep{lee2006complexity, lee2008departures} suggest that edge crossings and relations with shared arguments are common for PDTB. This distinction adds to the difficulty of exploring correlations of relations between the two frameworks, even if the two corpora are built on the same texts.  

In addition, in RST-DT, an inventory of 78 relations is used, which can be grouped into 16 classes. These relations can be divided into \textit{subject matter} relations (informational relations in \citet{moore-pollack-1992-problem}), which are relations whose intended effect is for readers to recognize them, and \textit{presentational} relations, which are intended to increase some inclination in readers~\citep{mann1988rhetorical} (intentional relations in \citet{moore-pollack-1992-problem}). For each relation, only one sense label can be attached. In contrast, PDTB adopts a three-level sense hierarchy, and more than one sense label can be annotated for a pair of arguments. As shown in the example, annotation 2 and annotation 3 are annotations for the same argument pair, but different sense labels are assigned. In previous studies that explore the alignment of RST and PDTB discourse relations, these cases typically require manual inspection to determine the closest matching PDTB relation to RST~\citep{demberg2019compatible}. Moreover, PDTB does not take intentional relations into account but focuses on semantic and pragmatic relations.

The combination of these factors makes it challenging to investigate the relationship between discourse relations of different annotation frameworks. Even in empirical studies that make use of corpora annotated in multiple frameworks in parallel, expert knowledge and manual examination are still required. To tackle the challenge caused by differences in discourse segmentation, \citet{demberg2019compatible} employ the strong nuclearity hypothesis~\citep{marcu2000theory}\footnote{A relation that holds between two spans should also hold between the nuclei of the two spans.} to facilitate the string matching process of aligning PDTB arguments and RST segments. While this method alleviates the limitation 
of exact string matching of arguments/EDUs, it relies on a corpus annotated with multiple frameworks in parallel. Furthermore, it is conceivable that the relations left out in their analysis because of violating the principle of strong nuclearity hypothesis are not necessarily irrelevant for the goal of enabling joint usage of RST and PDTB. 

In this study, we propose a fully automatic method for this task. We take inspiration from advances in label embedding techniques and an increasing body of research endeavors to harness label information in representation learning, such as supervised contrastive learning~\citep{khosla2020supervised, gunel2021supervised, suresh-ong-2021-negatives}. Instead of using string matching to identify the closest PDTB arguments and RST EDUs with the aim of discovering potentially analogous relations, we try to learn label embeddings of the relation inventories and compare the label embeddings. 

Our contributions can be summarized as follows:
\begin{itemize}
\item We propose a label embedding based approach for exploring correlations between relations of different discourse annotation frameworks. The method is fully automatic and eliminates the need of matching arguments of relations.
\item We conduct extensive experiments on different ways of encoding labels on RST-DT and PDTB 3.0. 
\item We develop a metric for evaluating the learnt label embeddings intrinsically and perform experiments to evaluate the method extrinsically.
\end{itemize}

\section {Related Work} \label{related-work}

\textbf{Mapping discourse relations} Existing research on mapping discourse relations of different frameworks can be categorized into three types~\citep{fu-2022-towards}: a. identifying a set of commonly used relations across various frameworks through analysis of definitions and examples~\citep{hovy1992parsimonious, bunt2016iso, benamara-taboada-2015-mapping}; b. introducing a set of fundamental concepts for analyzing relations across different frameworks~\citep{chiarcos-2014-towards, sanders2018unifying}; c. mapping discourse relations directly based on corpora annotated in multiple frameworks in parallel~\citep{rehbein-etal-2016-annotating}. The third approach is closer to our method, and we summarize studies in this direction here. ~\citet{rehbein-etal-2016-annotating} compare coherence relations of PDTB and CCR frameworks on the basis of a spoken corpus annotated in the two frameworks. They find that differences in annotation operationalisation and granularity of relation definition lead to many-to-many mappings.~\citet{demberg2019compatible} show similar findings when mapping relations of RST-DT and PDTB 2.0. To mitigate issues caused by segmentation differences, they use the \textit{strong nuclearity hypothesis}~\citep{marcu2000theory} so that relations that have greater scope than two adjacent EDUs can be covered in their studies. With this method,~\citet{costa-etal-2023-mapping} maps RST with PDTB 3.0. ~\citet{scheffler2016mapping} propose a method of mapping RST and PDTB relations on a German corpus annotated according to both frameworks. Explicit connectives in PDTB are used as anchors of relations, with some exceptions. It is found that 84.4\% of such PDTB explicit connectives can be matched to an RST relation. The results are not surprising, as phrases that begin with a strong discourse marker are specified as EDUs~\citep{carlson2001discourse}, and a relation is likely to be attached.
~\citet{stede-etal-2016-parallel} annotate a corpus with discourse information in RST and SDRT and argumentation information. A set of rules are applied to harmonize the segmentations, and structural transformation into a common dependency graph format is performed.~\citet{bourgonje-zolotarenko-2019-toward} try to induce PDTB implicit relations from RST annotation. Segmentation differences present a challenge, and even if the two annotations overlap in segmentation in some cases, different relations are annotated. This observation is consistent with~\citet{demberg2019compatible}.

\textbf{Label embeddings} Label embeddings have been proven to be useful in CV~\citep{7293699, NIPS2009_1543843a, zhang2022use} and NLP taks~\citep{wang-etal-2018-joint-embedding, zhang-etal-2018-multi, miyazaki-etal-2019-label}. Conventionally, one-hot encoding is used to represent labels, which suffers from three problems: lack of robustness to noisy labels~\citep{gunel2021supervised}, higher possibility of overfitting~\citep{sun2017label} and failure to capture semantic correlation of labels. Learning meaningful label representations is helpful for mitigating these problems and the semantics of labels can be used as additional information to improve model performance. It is shown that label embeddings are effective in data-imbalanced settings and zero-shot learning~\citep{zhang-etal-2022-label}.

Label embeddings can be representations from external sources, such as BERT~\citep{xiong-etal-2021-fusing}, or can be randomly initialized~\citep{zhang-etal-2022-label}. Another approach is to learn label embeddings during model training.~\citet{7293699} propose a method of learning label embeddings from label attributes while optimizing for a classification task.~\citet{wang-etal-2018-joint-embedding} introduce an attention mechanism that measures the compatibility of embeddings of input and labels. Additional information can be incorporated in learning label embeddings, such as label hierarchy~\citep{chatterjee-etal-2021-joint, zhang2022use, miyazaki-etal-2019-label} and textual description of labels~\citep{zhang2023description}.

\section{Method} \label{method}

\textbf{Problem statement} Given a corpus annotated in one discourse annotation framework $D_1=\{X_m, Y_m\}_{m=1}^{M}$ and another corpus annotated in a different annotation framework $D_2=\{X_n, Y_n\}_{n=1}^{N}$, where $X$ denotes input sequences formed by pairs of arguments, $X_i=$ $A^{(1)}_1$ $...$ $A^{(1)}_a$, $A^{(2)}_1$ $...$ $A^{(2)}_b$, and $Y$ represents relation label sets of the two frameworks, $Y_{D_1}$ = $\{y_1, y_2, ..., y_k\}$ and $Y_{D_2}$ = $\{y_1, y_2, ..., y_c\}$. The task is to learn a correlation matrix $\mathbf{\textit{R}}$ between $Y_{D_1}$ and $Y_{D_2}$, which is a $2d$ matrix of shape $k$ $\times$ $c$. Our method is to learn embeddings for members of $Y_{D_1}$ and $Y_{D_2}$ and the widely used cosine similarity can be used as a measure of distance between the embedding vectors. The label embedding learning method is the same for $D_1$ and $D_2$ and we use $D_1$ as an example in the following.

We apply the vanilla version of label-anchored contrastive learning in~\citet{zhang-etal-2022-label} as the backbone. For an input sequence $X_i$, we use a pre-trained language model as the input encoder $f_{InEnc}$. Without losing generality, we choose the popular~\textit{bert-base-uncased} model from the Huggingface transformers library~\citep{wolf-etal-2020-transformers}. For $X_i$ pre-processed as $X_i=$ $[CLS]$ $A^{(1)}_1$ $...$ $A^{(1)}_a$ $[SEP]$ $A^{(2)}_1$ $...$ $A^{(2)}_b$ $[SEP]$, 
the representation of the $[CLS]$ token is used as the representation of the input sequence:

\begin{equation} \label{input_encoder_eq}
\mathit{\mathbf{E}_{X_i} = f_{InEnc}({X_i})}       
\end{equation}

\noindent where the input sequence representation $\mathbf{E}_{X_i}$ is of shape $(a+b+3) \times dim$, where $dim$ is the dimension of the output from the language model and $a$ and $b$ are the maximum lengths that the arguments are padded to. We empirically find that removing the non-linear transformation to $\mathbf{E}_{X_i}$ in~\citet{zhang-etal-2022-label} yields better performance for our task.

We explored different options of label encoders, including:
adding a BERT model~\citep{devlin-etal-2019-bert} (${LbEncBert}$); using a RoBERTa model~\citep{liu2020roberta}, which is trained with the next sentence prediction objective removed (${LbEncRoberta}$); randomly initializing from a uniform distribution (${LbEncRand}$); adding text description of the labels (${LbEncDesc}$), where the label and the description are processed in the form $[CLS] label [SEP] description [SEP]$, and the representation of $[CLS]$ is used as the label representation; and adding sense hierarchy information, where we use the hierarchical contrastive loss proposed by~\citet{zhang2022use} and apply different penalty strengths to losses at different levels (${LbEncHier}$). As we use language models or trainable layers as label encoders, the label embeddings are learnable.

With a label encoder $g_{LbEnc}$, for $k$ total relations in $D_1$, we obtain a table $\mathbf{\textit{T}}$ of shape $k$$\times$$lbDim$, where $lbDim$ is the output dimension of the label encoder. Thus, for a label $y_{l_{l=1}^{k}}$, its label embedding vector $\mathbf{E}_{y_l}$ is the $({l-1})^\text{th}$ row of $\mathbf{\textit{T}}$.

\textbf{Instance-centered contrastive loss} We apply the method in~\citet{zhang-etal-2022-label} to compute the instance-centered contrastive loss $\mathcal{L}_{ICL}$:

\begin{equation}\label{instance-centered-loss}
\mathcal{L}_{ICL} = -\frac{1}{N} \sum\limits_{X_i, Y_i} \log \frac{  
e^{\Phi(\mathbf{E}_{X_i}, \mathbf{E}_{Y_i}) / \tau} }
{ \sum_{1\leq l \leq K} e^{\Phi(\mathbf{E}_{X_i},  \mathbf{E}_{Y_l}) / \tau} }          
\end{equation}

\noindent where $N$ denotes batch size, $X_i$ is an instance in a batch, and $Y_i$ is its label, $\Phi$ represents a distance metric between the representations of the input and label embeddings, and cosine similarity is used in the experiment. $\tau$ denotes the temperature hyper-parameter for scaling, and lower values of $\tau$ increase the influence of hard-to-separate examples in the learning process~\citep{zhang2021temperature}. By minimizing this loss, the distance between instance representations and the corresponding class label embeddings is reduced, resulting in label embeddings that are compatible with input representations.

\textbf{Label-centered contrastive loss} The purpose of this loss is to reduce the distance between instances that have the same labels. For a batch with a set of unique classes $C$, $c$ represents a member, $P_{c}$ denotes the set of instances in a batch that have the label $c$ and $N_{c}$ represent the set of negative examples for $c$. A member in $P_{c}$ is represented by $X_p$ and a member in $N_{c}$ is denoted by $X_n$. The label-centered contrastive loss $\mathcal{L}_{LCL}$ can be computed with:

\begin{equation} \label{label-centered-loss} 
\mathcal{L}_{LCL}= -\frac{1}{C}\sum\limits_{c \in C}\sum\limits_{X_p \in P_{c}}\log \frac{  
e^{\Phi(\mathbf{E}_{X_p}, \mathbf{E}_{c}) / \tau} }
{ \sum_{X_n \in N_{c}} e^{\Phi(\mathbf{E}_{X_n},  \mathbf{E}_{c}) / \tau} }          
\end{equation}

As indicated in~\citet{zhang-etal-2022-label}, $\mathcal{L}_{ICL}$ and $\mathcal{L}_{LCL}$ mitigate the small batch size issue encountered in other types of contrastive learning, which makes them suitable for scenarios with limited computational resources. 

We add the following two supervised losses in the training objective, which we find effective empirically. 

\textbf{Label-embedding-based cross-entropy loss} As shown in Equation~\ref{labelemb-softmax}, a softmax function is applied to the $k$ label embeddings in $\mathbf{\textit{T}}$, yielding a probability distribution over the $k$ classes:  
\begin{equation} \label{labelemb-softmax} 
 p({y_l}) = \frac{e^{\mathbf{E}_{y_l}}} {\sum_{l=1}^{K} e^{\mathbf{E}_{y_l}} }
\end{equation}

Let $t_{y_l}$ denote the categorical encoding of the target ${y_l}$. The cross-entropy loss of classification based on label embeddings, denoted by $\mathcal{L}_{LEC}$, can be obtained with Equation~\ref{labelemb-cls-loss}:

\begin{equation} \label{labelemb-cls-loss} 
\mathcal{L}_{LEC} = -\sum\limits_{l=1}^{K} t_{y_l} \log p({y_l})
\end{equation}

The purpose of adding this loss is to make the label embeddings better separated from each other. 

\textbf{Canonical multi-class cross-entropy loss} We add the canonical cross-entropy loss for multi-class classification with input representations:

\begin{equation} \label{input-cls-loss} 
 \mathcal{L}_{ICE}=-\sum\limits_{i=1}^{N}\sum\limits_{l=1}^{K} c^{i}_l
 \log p(c^{i}_l)         
\end{equation}

\noindent where $N$ is the batch size, $K$ is the total number of classes, and $p(c^{i}_l)$ is the probability predicted for a class $c$. With this loss, the input representations are learnt to be effective for the classification task.

The total loss is the sum of the four losses. During inference, only vector matching between the representation of an input sequence $\mathbf{E}_{X_i}$ and the $k$ learnt embeddings $\mathbf{E}_{y_l}$ is needed, with the cosine similarity as a distance metric, for instance.

\begin{equation} \label{inference-eq} 
 \hat{y} =\argmax \limits_{1\leq l\leq k}(\Phi(\mathbf{E}_{X_i}, \mathbf{E}_{y_l})) )        
\end{equation}

\textbf{Baseline for relation classification} We run the~\textit{BertForSequenceClassification} model from the Transformers library as the baseline for discourse relation classification, which is trained with cross-entropy loss only, i.e. Equation~\ref{input-cls-loss}.

\textbf{Baseline for label embedding learning} 
Label embeddings are generally used for improving performance in classification tasks in previous studies~\citep{wang-etal-2018-joint-embedding, zhang-etal-2018-multi, xiong-etal-2021-fusing, zhang-etal-2022-label}. To compare with a method targeted at learning good label embeddings, we implement a baseline method, which is a combination of Equation~\ref{labelemb-softmax} and~\ref{labelemb-cls-loss}, but a softmax function is applied over the cosine similarities of an input $\mathbf{E}_{X_i}$ and each label embedding $\mathbf{E}_{y_l}$ in $\mathbf{\textit{T}}$ here, similar to the approach adopted in~\citet{zhang-etal-2018-multi} and~\citet{wang-etal-2018-joint-embedding}.

\textbf{Metric} After the model training stage, as the representations of the input sequences have been learnt for the relation classification task, we can leverage the average of the representations of input sequences $X$ that belong to a class $y_l$ as a proxy for the class representation, denoted by $\mathbf{H}_{y_l}$:

\begin{equation} \label{class-rep-proxy} 
 \mathbf{H}_{y_l} = \frac{1}{C} \sum\limits_{i=1}^{C}\mathbf{E}_{X_i} 
\end{equation}
where $C$ represents the number of instances in $X$.

Due to inevitable data variance, the learnt label embeddings $\mathbf{E}_{y_l}$ for a class ${y_l}$ may not be the same as $\mathbf{H}_{y_l}$, but it should have a higher correlation with $\mathbf{H}_{y_l}$ than label embeddings of the other classes. Hence, we compute the correlation matrix $\mathbf{\textit{M}}$ between the $k$ learnt label embeddings $\mathbf{E}_{y_j}$ and the $k$ class representation proxies $\mathbf{H}_{y_i}$, where ${0\leq j, i \leq {k-1}}$, with cosine similarity as the metric of correlation: 

\begin{equation} \label{labelemb-classproxy-corr} 
 \mathbf{\textit{M}}_{ij} = \Phi(\mathbf{H}_{y_i},\mathbf{E}_{y_j})
\end{equation}

For each class representation proxy, we normalize its correlation scores with the $k$ learnt label embeddings to a range of [0, 1].
The average of values at the main diagonal of $\mathbf{\textit{M}}$ is adopted as an overall measure of the quality of the learnt label embeddings:

\begin{equation} \label{labelemb-m} 
\mathcal{LEQ} = \frac{1}{K}\sum\limits_{i=0}^{K-1}\widetilde{\mathbf{\textit{M}}}_{ii}
\end{equation}

\begin{figure}[htbp] 
\begin{center}
\includegraphics[
  width=0.25\textwidth,height=0.1\textheight, scale=0.95]
  {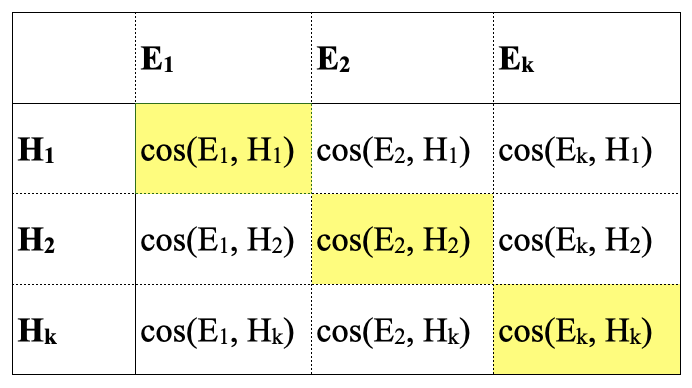} 
  \caption{Illustration of the correlation matrix $\mathbf{\textit{M}}$. $\mathbf{E}_{1...k}$ represents the $k$ learnt label embeddings and $\mathbf{H}_{1...k}$ denotes the $k$ class representation proxies. After normalization, the average of the values at the diagonal (colored) is the overall measure of the quality of the learnt label embeddings. }
  \label{label-emb-quality-viz}
\end{center}
\end{figure}

Figure~\ref{label-emb-quality-viz} shows the method of intrinsic quality estimation for learnt label emebeddings. 

\section{Experiments}

\subsection{Data Preprocessing} \label{preprocessing}
For the purpose of our research, it would be ideal to learn label embeddings for all the relations. However, the label embeddings are trained together with input representations in a multi-class classification task and data imbalance poses a challenge. Therefore, we focus on 16 relations for RST and PDTB L2 senses with more than 100 instances, following~\citet{kim-etal-2020-implicit}.

The RST trees in RST-DT are binarized based on the procedure in~\citet{ji-eisenstein-2014-representation} and the spans and relations are extracted. The 78 relations are mapped to 16 classes based on the processing step in~\citet{braud-etal-2016-multi}\footnote{\url{https://bitbucket.org/chloebt/discourse/src/master/preprocess_rst/code/src/relationSet.py}}. We take 20\% from the training set of RST-DT for validation purpose.  

For PDTB, we take sections 2-20 as the training set, sections 0-1 as the development set, and sections 21-22 as the test set, following~\citet{10.1162/tacl_a_00142}.

\subsection{Hyperparameters and Training}
We run each model three times with different random seeds and report the mean and standard deviation of the results. We use the AdamW optimizer~\citep{loshchilov2018decoupled} and clip L2 norm of gradients to 1.0. The learning rate is set to $1e-5$. The batch size is set to the maximum that the GPU device can accommodate. The total training epoch is set to 10 and we adopt early stop with patience of 6 on validation loss. 

The temperature $\tau$ for instance-centered contrastive loss and label-centered contrastive loss is set to 0.1. For the experiment with $LbEncHier$ label encoder, the penalty factor is $2^{1/2}$ for L1 loss and $2$ for L2 loss.

The learning rate for the baseline~\textit{BertForSequenceClassification} model is set to $5e-5$.

Our implementation is based on the PyTorch framework~\citep{paszke2019pytorch} and a single 12GB RTX3060 GPU is used for all the experiments.

\subsection{Results} \label{results}
Since we observe minimal discrepancies in data distributions between the training and test sets, we opt to utilize the test set for generating the class representation proxies necessary for the computation of the metric.

\begin{table*}[h] \scriptsize
\centering
\begin{tabular}{lllll}
\hline
\textbf{Data} & \textbf{Label enc.} & \textbf{Acc.} & \textbf{F1} & \textbf{Label emb.}\\
\hline
\multirow{6}{*}{PDTB total} & ${LbEncBert}$ & 69.45($\pm$ 0.18) & 57.80($\pm$ 0.85) & 93.84($\pm$ 0.37) \\ 
 & ${LbEncRoberta}$ & 69.34($\pm$ 0.46) &      58.10($\pm$ 0.15) & \underline{\textbf{94.23($\pm$ 0.74)}}  \\ 
 & ${LbEncRand}$ & \underline{\textbf{69.87($\pm$ 0.80)}} & \underline{\textbf{59.00($\pm$ 0.62)}} & 89.32($\pm$ 0.01) \\ 
 & ${LbEncDesc}$ & 69.16($\pm$ 0.26) & 57.53($\pm$ 0.14) & 93.58($\pm$ 0.42) \\ 
 & ${LbEncHier}$ & 69.21($\pm$ 0.45) &  56.70($\pm$ 0.14) & 93.67($\pm$ 0.23) \\ 
 & $Baseline$ & 69.42($\pm$ 0.46) &  58.73($\pm$ 0.78) & 79.15($\pm$ 2.06)   \\ \hline
\multirow{6}{*}{RST}
 & ${LbEncBert}$ & 64.62($\pm$ 0.90) & 44.86($\pm$ 1.85) & \underline{\textbf{78.64($\pm$ 1.02)}} \\ 
 & ${LbEncRoberta}$ & \underline{\textbf{65.20($\pm$ 0.07)}} & 
 45.39($\pm$ 0.60) & 76.56($\pm$ 0.85)  \\ 
 & ${LbEncRand}$ & 65.09($\pm$ 0.70) & 45.53($\pm$ 4.82) & 69.98($\pm$ 3.10) \\ 
 & ${LbEncDesc}$ & 64.62($\pm$ 0.21) & 43.69($\pm$ 1.20) & 74.18($\pm$ 0.91) \\ 
 & ${LbEncHier}$ & 63.66($\pm$ 0.50) &  41.30($\pm$ 0.39) & 74.54($\pm$ 0.77) \\ 
 & $Baseline$ & 63.55($\pm$ 0.23) & \underline{\textbf{48.57($\pm$ 0.73)}} & 48.21($\pm$ 1.27)   \\ \hline
\end{tabular}
\vspace{-1mm}
\caption{\label{PDTB-total-rst-results} 
With results over three runs collected, the Pearson correlation coefficient between classification accuracy and label embedding scores is 0.5814 and it is 0.8187 between f1 and label embedding scores, both with $p$ < 0.05), which shows that the learnt label embeddings are closely related to F1 scores.}
\end{table*}

Table~\ref{PDTB-total-rst-results} shows the experimental results for PDTB and RST. Explicit and implicit relations for PDTB are combined. After the preprocessing step, 16 relations remain for both PDTB and RST. 

It can be observed that the performance of label embedding learning on RST is lower than PDTB. 
Moreover, adding label embeddings generally lowers F1 compared with training with cross-entropy loss only. The decrease in F1 might be related to data sparsity when more learning objectives are added but the data amount is the same, which is visible when supplementary information of labels is added, as shown by cases of ${LbEncDesc}$ and ${LbEncHier}$. This phenomenon is rather pronounced for RST, which has a much smaller data amount. Additionally, although the label encoder ${LbEncRand}$ works best for the classification task, the learnt label embeddings rank the lowest among the different options. Through examination, we find that with this approach, the label embeddings of different classes are not close to the class representation proxies and we conjecture that during training, the label embeddings are mainly used as anchors, as in~\citet{zhang-etal-2022-label}, but the input representations are better learnt, hence the higher classification accuracy and F1 score.~\citet{zhang-etal-2022-label} did not report other options of label encoders than random initialization and their focus is classification accuracy.

\subsection{Data Augmentation for RST} \label{rst-augmentation}
To improve the performance on RST, we use back translation as a means of data augmentation. We translate all the files containing EDUs in the training set (only) into French and translate the French texts back into English, using Google Translate\footnote{\url{https://translate.google.com/}}. Data augmentation is not performed for~\textit{Elaboration} and~\textit{Joint}, which are the two largest classes in RST-DT, to achieve a more balanced data distribution.

Based on the results shown in Table~\ref{PDTB-total-rst-results}, we choose ${LbEncRoberta}$ in the following experiments because of its good performance but results with ${LbEncBert}$ are comparable.

Table~\ref{rst-data-aug} shows the results. The F1 scores and label embedding scores are improved to a large margin. As back translation is performed at the EDU level, it is unavoidable that errors are introduced, and given that data augmentation is not performed for the two largest classes, their influence on the results is reduced, hence the lower classification accuracy.

\begin{table}[H] \scriptsize
\centering
\begin{tabularx}{0.9\columnwidth}{cccc}
\hline  
 &\textbf{Acc.} & \textbf{F1} & \textbf{Label emb.} \\ 
 \hline
 +aug. & 62.75($\pm$ 0.79) & 50.76($\pm$ 0.94) & 92.96($\pm$ 0.90) \\ 
 -aug. & 65.20($\pm$ 0.07) & 45.39($\pm$ 0.60) & 76.56($\pm$ 0.85) \\ 
\hline
\end{tabularx}
\vspace{-1mm}
\caption{\label{rst-data-aug} Results for RST with data augmentation (+aug) and without data augmentation (-aug).}
\end{table}

\begin{figure*}[h]
\centering
 \begin{subfigure}{\textwidth}
\hspace*{-3mm} \includegraphics[width=65mm, height=1.\columnwidth, scale=2.0,  angle=-90]{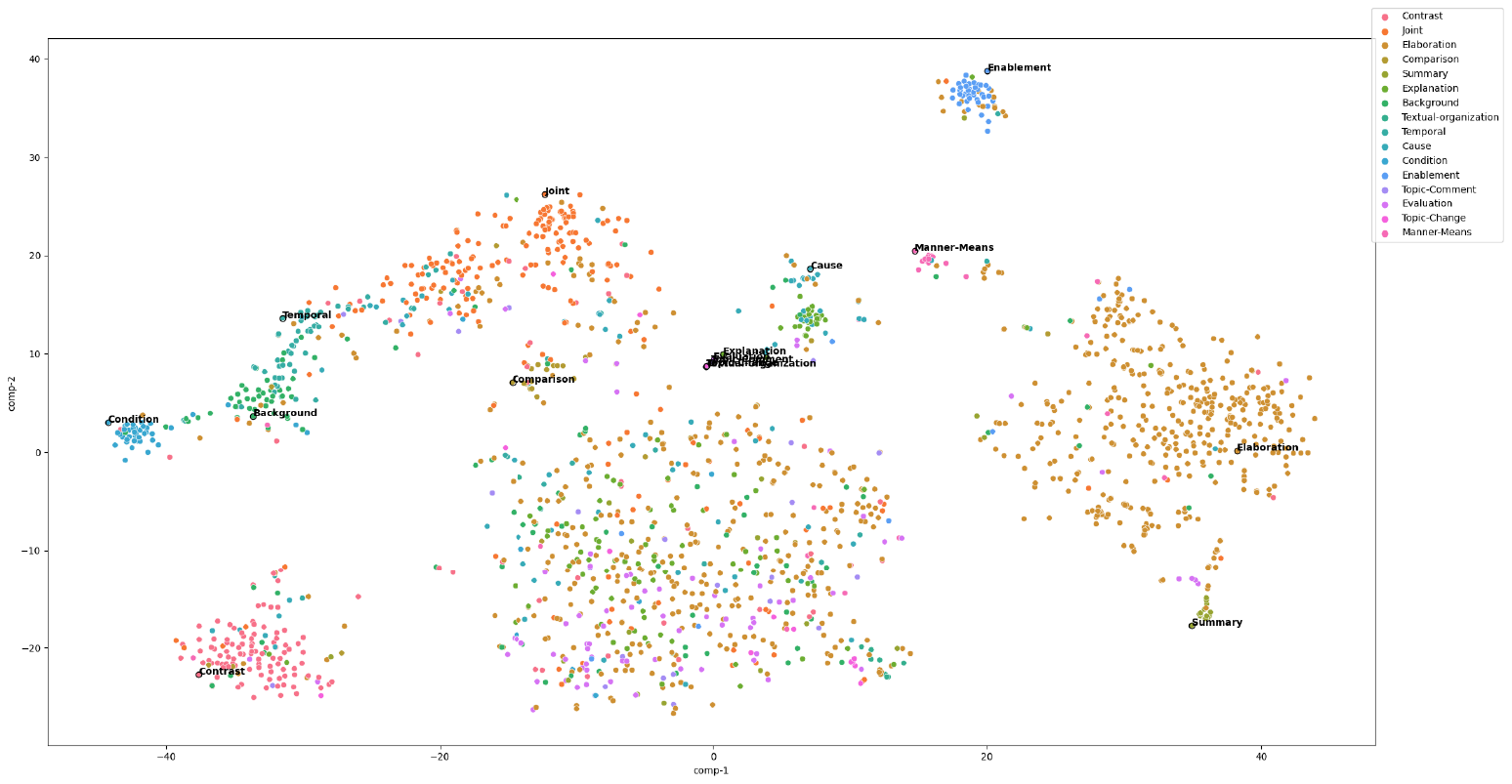}
\vspace{-1cm}
   \caption{}
   \label{rst-aug-roberta} 
\end{subfigure}
\begin{subfigure}{\textwidth}
  \hspace*{-3mm} \includegraphics[width=70mm, height=1.\columnwidth, scale=1.0, angle=-90]{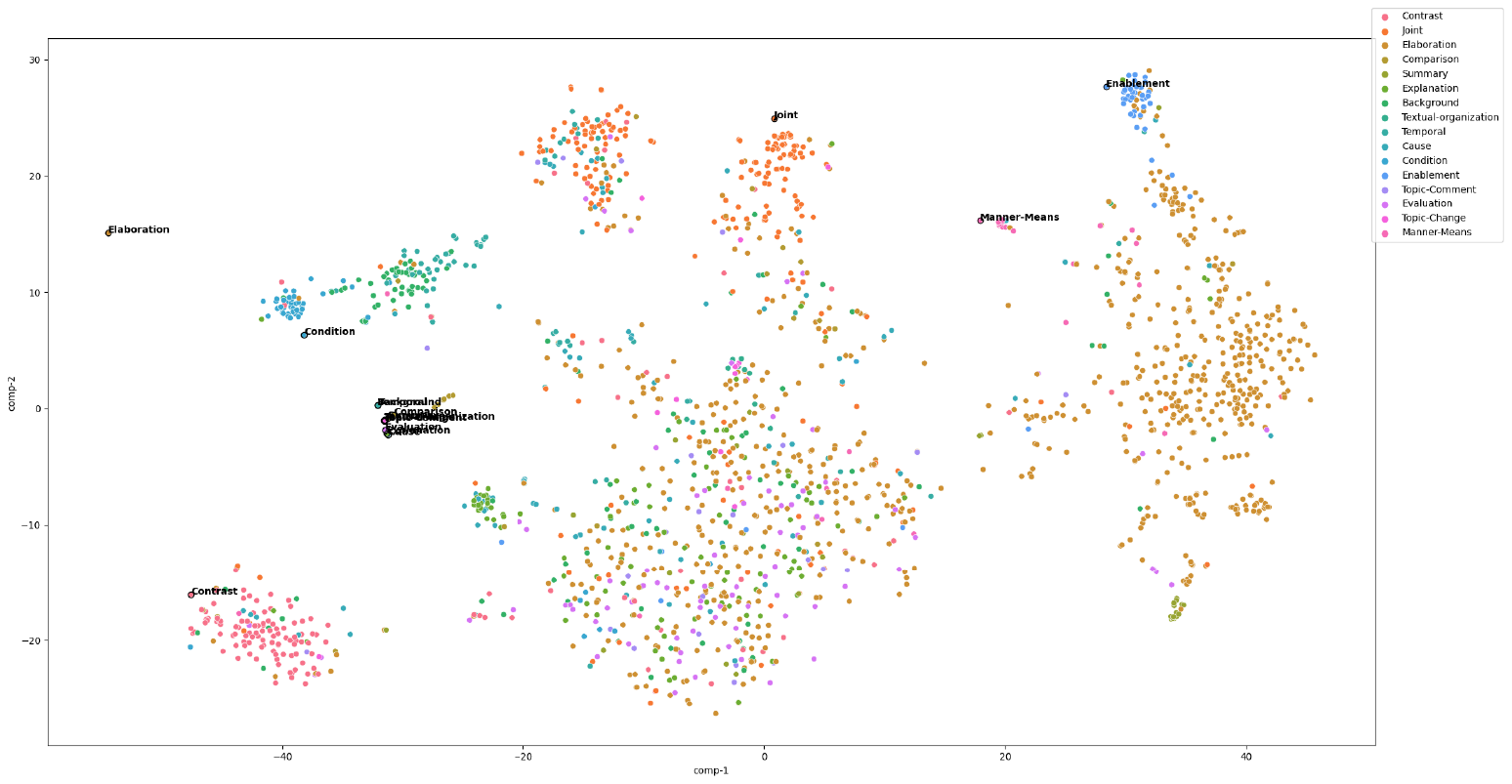}
  \vspace{-1cm}
   \caption{}
   \label{rst-no-aug}
\end{subfigure}
\caption[rst-lb-aug]{(a) Label embeddings learnt with data augmentation. (b) Label embeddings learnt without data augmentation. For visualization, we choose the label embeddings with the highest score from the three runs.} \label{rst-lb-overal}
\end{figure*}

Figure~\ref{rst-lb-overal} shows the T-SNE visualization plots of learnt label embeddings together with the class representation proxies for the test set of RST-DT. The label embeddings learnt with data augmentation are shown in Figure~\ref{rst-aug-roberta} in comparison with Figure~\ref{rst-no-aug}, where no data augmentation is performed. It is visible that in Figure~\ref{rst-aug-roberta}, more label embeddings fit into the class representation proxies while in Figure~\ref{rst-no-aug}, label embeddings of only six classes are close to the class representation proxies, and the rest form a nebula, which suggests that the label embeddings cannot be distinguished clearly from each other. In Figure~\ref{rst-aug-roberta}, label embeddings for five relations including \textit{Explanation}, \textit{Textual-Organization}, \textit{Topic-Comment}, \textit{Evaluation} and \textit{Topic-Change} show such behavior. \textit{Textual-Organization}, \textit{Topic-Comment}, and \textit{Topic-Change} are classes with a small amount of data and it is difficult to obtain good performance on these classes in a classification task. The reason for \textit{Explanation} and 
\textit{Evaluation} is not clear, and we leave it to future work. 

\subsection{Separate Experiments on PDTB Explicit and Implicit Relations} \label{pdtb-separate}
Previous studies~\citep{demberg2019compatible, sanders2018unifying} indicate that it is much easier to obtain consistent results on aligning PDTB explicit relations with relations from the other frameworks, while implicit relations are generally ambiguous and the consistency is much lower. Therefore, we conducted experiments on PDTB explicit and implicit relations separately. We use ${LbEncRoberta}$ in the experiments. After the data preprocessing step outlined in section~\ref{preprocessing}, 12 explicit relations and 14 implicit relations remain in the experiments. 

\begin{table}[h] \scriptsize
\centering
\begin{tabularx}{0.9\columnwidth}{llll}
\hline
\textbf{Data} & \textbf{Acc.} & \textbf{F1} & \textbf{Label emb.}\\
\hline
explicit & 88.98($\pm$ 0.41) & 79.19($\pm$ 0.64) & 99.15($\pm$ 0.60) \\ 
implicit & 56.05($\pm$ 0.56) &  40.56($\pm$ 0.81) & 82.21($\pm$ 0.85)  \\  \hline
\end{tabularx}
\vspace{-2mm}
\caption{\label{PDTB-expl-impl-results} Results of experiments on PDTB explicit relations and implicit relations.}
\end{table}

The classification results and label embedding learning results indicate that the learnt label embeddings for PDTB explicit relations are representative of the classes while the performance on implicit relations is sub-optimal. 

\subsection{Ablation Study}
We choose ${LbEncRoberta}$ and conduct ablation studies with PDTB explicit and implicit relations combined, similar to the experimental settings in Table~\ref{PDTB-total-rst-results}. The impact of each loss can be seen in Table~\ref{ablation-studies}.  

\begin{table}[h!] \scriptsize
\centering
\begin{tabularx}{0.92\columnwidth}{llll}
\hline
\textbf{Loss} & \textbf{Acc.} & \textbf{F1} & \textbf{Label emb.}\\
\hline
-$\mathcal{L}_{ICL}$ & 68.22($\pm$ 0.44) & 53.65($\pm$ 1.13) & 91.36($\pm$ 0.73) \\ 
-$\mathcal{L}_{LCL}$ & 65.02($\pm$ 0.47) &  51.23($\pm$ 1.62) & 80.37($\pm$ 1.42)  \\  
-$\mathcal{L}_{LEC}$ & 69.32($\pm$ 0.30) &  57.57($\pm$ 0.87) & \textbf{94.36($\pm$ 0.37)}  \\ 
-$\mathcal{L}_{ICE}$ & \textbf{69.88($\pm$ 0.09)} &  56.94($\pm$ 0.36) & 90.79($\pm$ 0.76)  \\  
$Total$ &  69.34($\pm$ 0.46) &  \textbf{58.10($\pm$ 0.15)} & 94.23($\pm$ 0.74)  \\  
\hline
\end{tabularx}
\vspace{-2mm}
\caption{\label{ablation-studies} Effect of each loss on model performance.}
\end{table}
As shown, the label-centered contrastive loss ($\mathcal{L}_{LCL}$) is of paramount importance for the model's performance, followed by the instance-centered contrastive loss ($\mathcal{L}_{ICL}$) and canonical cross-entropy loss ($\mathcal{L}_{ICE}$). This differs from the findings in~\citet{zhang-etal-2022-label}, where $\mathcal{L}_{ICL}$ is the primary contributing factor to their results, indicating the distinct nature of our respective tasks.
$\mathcal{L}_{LEC}$ has some effect on F1 score of the classification task. 

\section{RST-PDTB Relation Mapping}\label{mapping}
\subsection{Mapping Results}

Table~\ref{mapping-result} shows the results of mapping 11 RST relations, with the five relations discussed in section~\ref{rst-augmentation} excluded, and 12 PDTB explicit relations discussed in section~\ref{pdtb-separate}. Two relations with highest values in cosine similarity (greater than 0.10) are presented.

\begin{table}[!ht] \tiny
\centering
\begin{minipage}[t]{0.48\linewidth}\centering
\begin{tabular}{|l|l|}
\hline
 \textbf{RST} & \textbf{Relations in PDTB} \\
\hline
contrast & \makecell[l]{concession(0.25),\\ contrast(0.24)} 
\\  \hline
manner-means & \makecell[l]{manner(0.30),\\ purpose(0.25)}
\\   \hline
cause & \makecell[l]{cause(0.40), \\level-of-detail(0.17)}
\\   \hline
 background & \makecell[l]{synchronous(0.23),\\ manner(0.16)} 
 \\   \hline
condition & \makecell[l]{condition(0.39),\\ purpose(0.18)}
\\   \hline
  elaboration & \makecell[l]{concession(0.19),\\ disjunction(0.14)}
  \\   \hline
 enablement & \makecell[l]{manner(0.24),\\ purpose(0.18)}
 \\  \hline
 summary & \makecell[l]{contrast(0.35), \\level-of-detail(0.23)}
 \\  \hline
 joint & \makecell[l]{disjunction(0.25),\\ synchronous(0.20)}
 \\   \hline
 temporal & \makecell[l]{asynchronous(0.24), \\purpose(0.20)}
 \\  \hline
 comparison & \makecell[l]{purpose(0.17), \\level-of-detail(0.16)}
 \\   \hline
\end{tabular}
 \end{minipage}
 \hfill%
\begin{minipage}[t]{0.48\linewidth}\centering
\begin{tabular}{|l|l|}
\hline
 \textbf{PDTB} & \textbf{Relations in RST}\\   \hline
 conjunction & \makecell[l]{contrast(0.22),\\ elaboration(0.13)} \\   \hline
concession  & \makecell[l]{contrast(0.25), \\elaboration(0.19)}
\\   \hline
cause  & \makecell[l]{cause(0.40),\\ manner-means(0.20)}\\ \hline
level-of-detail & \makecell[l]{manner-means(0.25), \\summary(0.23)}
\\ \hline
synchronous & \makecell[l]{background(0.23), \\joint(0.20)}
\\ \hline
disjunction &  \makecell[l]{joint(0.25), \\temporal (0.16)}
\\ \hline
manner  &  \makecell[l]{manner-means(0.30),\\ enablement(0.24)}
\\  \hline
condition  &  \makecell[l]{condition(0.39),\\ summary(0.15)}
\\ \hline
substitution & \makecell[l]{manner-means(0.17),\\ summary(0.17)}
\\ \hline
asynchronous & \makecell[l]{temporal(0.24), \\joint(0.19)}
\\ \hline
contrast & \makecell[l]{summary(0.35),\\ background(0.13)}\\ \hline
purpose & \makecell[l]{manner-means(0.25),\\ temporal(0.20)}
\\ \hline
\end{tabular}
\end{minipage}
\caption{\label{mapping-result} Mapping between 11 RST relations and 12 PDTB explicit relations. The values in brackets represent the cosine similarity scores. 
}
\end{table}

The table on the left shows the mapping results from RST's perspective. For most of the RST relations, a PDTB relation can be identified as having a much higher value ($\ge$ 0.40) than the others.

The table on the right shows the mapping results from PDTB's perspective. As relation distributions are different, it is understandable that the two perspectives are not symmetric. 

\subsection{Extrinsic Evaluation}
We compare our results with those provided by~\citet{costa-etal-2023-mapping}, where the approach proposed in~\citet{demberg2019compatible} is adopted but results are updated to PDTB 3.0.
As shown in section~\ref{pdtb-separate}, label embeddings learnt for PDTB explicit relations are more reliable and we choose to focus on the mapping between PDTB explicit relations and RST relations. Based on Table~\ref{mapping-result}, we exclude PDTB's \textit{Substitution} relation in the experiments, for which no RST relations with higher similarity are observed, and relabel 11 PDTB explicit relations with RST labels based on Table~\ref{pdtbexpl-relabeled}. 

While we choose the RST label mostly based on cosine similarity shown in Table~\ref{mapping-result}, we take distribution of relations into account. For example, PDTB's \textit{Conjunction} relation is not mapped to RST's \textit{Contrast} relation but to \textit{Elaboration}, because \textit{Conjunction} is a large class in PDTB, similar to \textit{Elaboration} in RST, and relabelling in this way may keep the label distribution balanced. Meanwhile, in our preliminary experiments, mapping PDTB's \textit{Contrast} relation to RST's \textit{Summary} relation yields poor performance. Therefore, we relabel PDTB's \textit{Contrast} as RST's \textit{Contrast} relation based on the results from RST's perspective.

Similarly, we relabel PDTB explicit relations based on the results shown in~\citet{costa-etal-2023-mapping}\footnote{Table 5 in their paper.}. As their results are a mapping of 12 fine-grained RST relations and seven L2 PDTB relations, a direct mapping comparable to ours is not available. Thus, for a PDTB relation, if there are multiple mapped RST relations that fall under a broad class, the corresponding RST relation from the 16 categories is chosen, and the average of the percentages for the mapped classes is taken as the mapping strength, similar to cosine similarity in our results. For instance, PDTB \textit{Concession} is mapped to \textit{Contrast} (61.0\%), \textit{Antithesis} (84.0\%), and \textit{Concession} (88.0\%), which are fine-grained relations under RST \textit{Contrast}, and the mapping strength is the average of the three percentages, i.e., 0.78.

\begin{table}[!ht] \scriptsize
\centering
\begin{tabular}{|l|l|l|}
\hline
 \makecell[l]{\textbf{Original PDTB}\\ \textemdash Sense Labels}  & \makecell[l]{\textbf{RST Labels}\\ \textemdash Our method} & \makecell[l]{\textbf{RST Labels}\\ \textemdash\citet{costa-etal-2023-mapping}} \\
\hline
 concession & contrast (0.25) &  contrast (0.78) \\  
\hline
 contrast & contrast (0.24)  & contrast (0.26) \\ 
\hline
conjunction & elaboration (0.13) & joint (0.84) \\ 
\hline
 manner & manner-means (0.30) &  \textemdash \\
\hline
cause & cause (0.40) & explanation (0.69) \\ 
\hline
  synchronous & background (0.23) & temporal (0.98)\\ 
\hline
 condition & condition (0.39) & condition (0.84) \\  
\hline
 disjunction & joint (0.25) & \textemdash \\
\hline
 asynchronous & temporal (0.24) & temporal (0.94)  \\ 
\hline
 level-of-detail & manner-means (0.25) & \textemdash  \\ 
 \hline
 purpose & manner-means (0.25) & \textemdash \\ 
\hline
\end{tabular}
\caption{\label{pdtbexpl-relabeled} Relabelling of PDTB explicit relations. The similarity scores are shown in brackets.}
\end{table}

Based on our alignment results, 14964 instances of PDTB explicit relations are relabeled, and with the result in~\citet{costa-etal-2023-mapping}, 13905 PDTB instances are relabeled. Adding PDTB data to RST data causes a marked performance drop. The best result is obtained with an ensemble model, which is formed by stacking a model trained with a target of minimizing supervised contrastive loss, a model trained to minimize a label embedding loss, the label embeddings being randomly initialized, and a model that takes the input for relation classification. The output distributions of the three models are averaged and used for model prediction, and a cross-entropy loss is to be reduced in addition to the supervised contrastive loss and label embedding loss. As shown in Table~\ref{extrinsic-eval}, the performance with our method is slightly higher. 

\begin{table}[!ht] \scriptsize
\centering
\begin{tabular}{l|l|l}
\hline
  & Acc. & F1 \\
\hline
\citet{costa-etal-2023-mapping} & 62.13 $\pm$ 0.34    &  46.96$\pm$0.43   \\ 
\hline
Our method & 63.13 $\pm$ 1.12   &  47.95$\pm$ 1.07    \\ 
\hline
-PDTB aug. & 63.82$\pm$ 1.07 & 48.72$\pm$ 0.11 
\\  \hline
\end{tabular}
\caption{\label{extrinsic-eval} Results of extrinsic evaluation.}
\end{table}

\section {Conclusions}
We propose a method of automatically aligning discourse relations from different frameworks. By employing label embeddings that are learned concurrently with input representations during a classification task, we are able to circumvent the challenges posed by segmentation differences, a significant hurdle encountered in prior studies. We perform intrinsic and extrinsic evaluation of the results of the method. 
Similar to other empirical studies, our method is affected by the amount of data, and we have to exclude some relations for which there may be too little training data to learn reliable label embeddings. A comparison with a theoretical proposal, such as ISO 24617-8 \citep{prasad-bunt-2015-semantic}, merits investigation in future work. The method may extend beyond labelling of discourse relations to alignment of any label sets, leaving the possibility of application to a variety of scenarios.\footnote{We thank the anonymous reviewers for pointing out the two directions.} 

\section{Acknowledgments}
We thank the anonymous reviewers for insightful
feedback and suggestions. Our thanks also go to Mark-Jan Nederhof for discussions and Craig Myles for the suggestion of using the diagonal entries of the normalized correlation matrix as a metric.

\section{Bibliographical References}\label{sec:reference}
\bibliographystyle{lrec-coling2024-natbib}
\bibliography{lrec-coling2024-example}

\section{Language Resource References}

\begin{hangparas}{0.35cm}{1}
Lynn Carlson, Daniel Marcu, and Mary Ellen Okurowski. 2002. \textit{RST Discourse Treebank}. distributed
via LDC. Philadelphia: Linguistic Data Consortium: LDC2002T07, Text resources, 1.0, ISLRN:	299-735-991-930-2. 
\\
\end{hangparas}

\begin{hangparas}{0.35cm}{1}
Rashmi Prasad, Bonnie Webber, Alan Lee, and Aravind Joshi. 2019. \textit{Penn Discourse Treebank Version 3.0}.
LDC. distributed via LDC. Philadelphia: Linguistic Data Consortium:
LDC2019T05, Text resources, 3.0, ISLRN 977-
491-842-427-0.
\end{hangparas}

\label{lr:ref}
\bibliographystylelanguageresource{lrec-coling2024-natbib}
\bibliographylanguageresource{languageresource}

\end{document}